# Binary classification for perceived quality of headlines and links on worldwide news websites, 2018-2024

Austin McCutcheon
Department of Computer Science
Lakehead University
Orillia, Canada
aomccutc@lakeheadu.ca

Thiago E. A. de Oliveira
Department of Computer Science
Lakehead University
Orillia, Canada
talvesd@lakeheadu.ca

Aleksandr Zheleznov
Department of Computer Science
Lakehead University
Orillia, Canada
azhelezn@lakeheadu.ca

Chris Brogly
Department of Computer Science
Lakehead University
Orillia, Canada
cbrogly@lakeheadu.ca

*Abstract*—The proliferation of online news enables potential widespread publication of perceived low-quality news headlines/links. As a result, we investigated whether it was possible to automatically distinguish perceived lower-quality news headlines/links from perceived higher-quality headlines/links. We evaluated twelve machine learning models on a binary, balanced dataset of 57,544,214 worldwide news website links/headings from 2018-2024 (28,772,107 per class) with 115 extracted linguistic features. Binary labels for each text were derived from scores based on expert consensus regarding the respective news domain quality. Traditional ensemble methods, particularly the bagging classifier, had strong performance (88.1% accuracy, 88.3% F1, 80/20 train/test split). Fine-tuned DistilBERT achieved the highest accuracy (90.3%, 80/20 train/test split) but required more training time. The results suggest that both NLP features with traditional classifiers and deep learning models can effectively differentiate perceived news headline/link quality, with some trade-off between predictive performance and train time.

## I. Introduction

The massive output of online news content from many sources far exceeds manual human review capabilities for perceived quality, potentially enabling the proliferation of deceptive or misleading text such as clickbait or other perceived lower-quality texts. This perceived lower-quality information might also be sometimes be interpreted as fact. However, some content with lower journalistic standards, such as satire or political commentary, serves important cultural and social roles [1]. Complete censorship or removal of perceived lower-quality content is impractical, making optional quality classification and prediction a more balanced approach.

News quality classification models could reveal perceived lower-quality content and perceived higher-quality content, by providing users with statistical assessments given an established benchmark for content quality on news websites via ratings. Machine learning and natural language processing approaches for detecting news quality have been studied [2][3][4] but the scale of data used in existing work is generally limited to smaller data for texts. Focus is often placed on narrow classification tasks such as fake news detection or bias identification, with less attention to comprehensive quality assessment using a large-scale dataset [5][6].

As a result, we were interested in investigating a classification approach based on domain quality ratings utilizing traditional machine learning models and deep learning applied to a large-scale dataset. News domain quality ratings were initially computed in [7], where the first principal component of a PCA performed on an imputed domain quality dataset [7] was used as the basis for text quality assessment here. We trained models on a dataset of approximately 57 million URL links with 115 NLP features, examining PC1 values to decide binary classification labels. The dataset originally contained 196 linguistic features in total, however not all features were used. Additionally, we evaluated a fine-tuned DistilBERT model operating directly on link text rather than extracted features.

Our analysis reveals that ensemble methods, especially the bagging classifier, achieved strong performance with this worldwide news headline/link large dataset. The DistilBERT model obtained the highest accuracy but required significantly longer training time for the marginal increase in accuracy.

## II. Methods

### A. Data Collection

A large-scale dataset was constructed from the Common Crawl database, which contained 398,763,321 rows of news headline/link content. The data was sourced exclusively from .com domains and processed using an NLP features detector to generate 196 features for each row, along with the URL, link text, and article release time. The features spanned five categories: part-of-speech tags, Penn Treebank tags, syntactic dependencies, and named entity recognition. For more information on these categories, see [8]. One additional category included 21 additional numeric NLP measures such as word count and average length of bigrams/trigrams. The full list of these was omitted here for brevity, as some were disused after feature selection.

Quality labels were assigned using PC1 scores. PC1 overall is a quality measure between 0.0 and 1.0 that ranked websites based on relevant items such as factual accuracy, bias assessment, and review scores [7]. The PC1 methodology provided multiple expert consensus on news domain quality, so it is well-suited for supervised learning applications. A CSV file containing 614 unique domains with corresponding

PC1 values was used to assign quality labels to each individual text in our dataset based on the domain the text came from.

Domain extraction was performed by cleaning page URLs to the format "example.com" and matching them against the PC1 domain list. When a match was found, the corresponding PC1 value was assigned to all rows from that domain. This domain-level labeling approach assumed quality was consistent for content from the same source.

For binary classification, PC1 values were converted using the median threshold of 0.8163. Domains with PC1 scores above this threshold were labeled as high quality (1), while those below were labeled as low quality (0). The median threshold was selected to create a balanced dataset while maintaining a meaningful quality distinction (above 0.8).

The labeled dataset was exported from the database as CSV files, split by year due to memory constraints. All time periods were represented in model training. The dataset contained 66,803,765 entries with non-null PC1 values given domain matches spanning 2018-2024, although this was reduced down to 57,544,214 samples from class balancing (28,772,107 samples for each binary class). For DistilBERT training, the same dataset was used with link caption text replacing the NLP feature scores used in traditional supervised learning. Full information can be found in Table 1.

TABLE I: DATASET OVERVIEW

| Dataset Item | Value | Description |
| --- | --- | --- |
| Total Raw Entries (all >= 3 words) | 398,763,321 | Full dataset size from original Common Crawl source |
| Filtered by Domain with PC1 | 66,803,765 | Entries matched with PC1-scored domains |
| Final Balanced Dataset | 57,544,214 | After class balancing using random undersampling |
| Number of total Features | 196 | NLP-derived features from feature detector |
| Features after sparsity reduction | 115 | Features with >1% non-zero values retained |
| Time span | 2018-2024 | Article publication years |
| Unique labeled News Domains | 614 | Domains with Assigned PC1 scores |
| Label Type | Binary PC1 > 0.8 | Derived from continuous PC1 scores |
| Class distribution | 50%/50% | Final Class distribution after preprocessing |
| Train-Test Split | 80%/20% | Train-Test split used for all training |

### B. Preprocessing

Sparsity analysis was performed to identify and remove uninformative features for traditional supervised learning from the detector columns. Early experiments revealed that 81 of our 196 features were very sparse, which was slowing model training. For practicality the decision to drop any feature appearing in less than 1% of samples was made. This process eliminated 81 features, reducing the total from 196 to 115 features. The remaining features were converted to numpy arrays for model input.

Class imbalance was addressed through yearly stratified undersampling. Each yearly partition was analyzed individually, and undersampling was performed using the minority class count per year to ensure balance. A fixed random seed of 42 was applied to ensure reproducibility across experiments. This balancing procedure finally yielded a final dataset of 57,544,214 entries with an equal 50/50 class distribution.

The balanced dataset was partitioned using stratified sampling with an 80-20 train-test split, maintaining proportional representation of both quality classes in training and testing sets. Feature normalization was applied using scikit-learn's StandardScaler.

### C. Models

In total twelve models were evaluated: eleven traditional and one deep learning as shown in Table 2.

TABLE II: MACHINE LEARNING MODELS & PARAMETERS

| Model | Parameters |
| --- | --- |
| DistilBERT | batch_size=1536, epochs=5, learning_rate=2e-5, Weight_decay=0.01, fp16=True, optim="adamw_torch_fused" |
| Bagging Classifier | n_estimators=25, max_samples=0.6, max_features=0.6, n_jobs=3, random_state=42 |
| Multilayer Perceptron (MLP) Large | hidden_layer_sizes=(256, 128), max_iter=1000, learning_rate_init=0.001, early_stopping=True, validation_fraction=0.1, n_iter_no_change=10, alpha=0.001, random_state=42 |
| Multilayer Perceptron (MLP) Small | hidden_layer_sizes=(64, 32), max_iter=300, learning_rate_init=0.001, early_stopping=True, validation_fraction=0.1, n_iter_no_change=10, alpha=0.001, random_state=42 |
| Histogram-based Gradient Descent (HistGB) | max_depth=None, random_state=42 |
| Random Forest - Depth 30 | n_estimators=200, max_depth=30, n_jobs=-1, random_state=42 |
| Random Forest - Depth 15 | n_estimators=200, max_depth=15, n_jobs=-1, random_state=42 |
| Random Forest - Depth 8 | n_estimators=200, max_depth=8, n_jobs=-1, random_state=42 |
| Stochastic Gradient Descent (SGD) SVM | loss='hinge', max_iter=1000, tol=1e-3, n_jobs=-1, random_state=42 |
| Voting Classifier | Estimators: GaussianNB() SGDClassifier( loss='hinge', ax_iter=1000, tol=1e-3, random_state=42) RandomForestClassifier(n_estimators=25, max_depth=5, n_jobs=1, |

|  | random_state=42)<br>Voting: voting='hard', n_jobs=-1 |
|---|---|
| Gaussian Naïve Bayes | Default |
| Dummy Classifier (Stratified) | strategy="stratified", random_state=42 |

## D. Performance Metrics

Model performance was evaluated using multiple classification metrics. Accuracy measured the percentage of correct predictions across all test set samples. Precision measured the proportion of correct positive predictions among all positive predictions, while recall measured the proportion of actual positives correctly identified by the model. F1 scores were computed for all models, providing the harmonic mean of precision and recall. ROC AUC values were calculated for eleven of the twelve models, with the voting classifier excluded due to implementation constraints in the ensemble voting mechanism.

## III. RESULTS

The twelve models were trained and evaluated on a Windows 11 machine equipped with an AMD Ryzen 9 5900X 12-core processor (24 threads), 128 GB of DDR4 RAM at 2400 MHz, an NVIDIA GeForce 4090 GPU, and SSDs. Our findings suggest that both machine learning and deep learning can work well for distinguishing news quality given this dataset, with accuracy ranging from 88.1% for traditional ensemble methods to 90.3% for a deep learning model as shown in Table 3 below.

TABLE III: MACHINE LEARNING/DEEP LEARNING MODEL RESULTS

| Model Name | Train time (sec) | Accuracy | F1 Score | Precision | Recall | ROC AUC |
|---|---|---|---|---|---|---|
| DistilBERT Finetune | 51518 | 0.9027 | 0.9026 | 0.90 | 0.90 | 0.9748 |
| Bagging Classifier | 8191 | 0.8807 | 0.8831 | 0.88 | 0.88 | 0.9638 |
| Random Forest (200 estimators, Max Depth 30) | 8863 | 0.8733 | 0.8746 | 0.87 | 0.87 | 0.9050 |
| Multilayer Perceptron (Large) | 20021 | 0.8027 | 0.8073 | 0.80 | 0.80 | 0.8996 |
| Multilayer Perceptron (Small) | 4337 | 0.7430 | 0.7452 | 0.74 | 0.74 | 0.8399 |
| Random Forest (200 estimators, Max depth 15) | 6026 | 0.7396 | 0.7427 | 0.74 | 0.74 | 0.8214 |
| HistGB | 453 | 0.6795 | 0.6820 | 0.68 | 0.68 | 0.7511 |
| Random Forest (200 estimators, Max depth 8) | 3533 | 0.6037 | 0.6510 | 0.61 | 0.60 | 0.6696 |
| SGD SVM | 157 | 0.5627 | 0.6217 | 0.57 | 0.56 | 0.5930 |
| Voting Classifier | 1498 | 0.5602 | 0.6228 | 0.57 | 0.56 | N/A |
| Gaussian NB | 31.25 | 0.5392 | 0.6068 | 0.54 | 0.54 | 0.5590 |
| Dummy Stratified | 1.38 | 0.4998 | 0.4998 | 0.50 | 0.50 | 0.4998 |

As anticipated, our stratified dummy classifier achieved about 50% accuracy. Traditional models included Gaussian Naïve Bayes, SGD SVM, and Random Forests with varying depths. Neural networks included Multilayer Perceptrons of different capacities. Ensemble methods tested were Bagging and Voting classifiers. One deep learning model, DistilBERT, was fine-tuned for comparison. We interpret the results more in the Discussion section.

Additionally, cross-validation was also performed to assess model generalization beyond the single train-test split. Due to memory constraints and processing time associated with the large dataset size, cross-validation was applied only to the highest-performing traditional machine learning model; we argue the high sample size offsets the need for this on every model. Five-fold stratified cross-validation was implemented, with each fold maintaining the balanced class distribution. For each fold, the model was cloned to prevent interference between evaluations, as memory limitations required processing only one-fold at a time.

TABLE IV: 5-FOLD CROSS VALIDATION RESULTS FOR BAGGING CLASSIFIER

| 5-Fold Bagging | Accuracy | F1 Score | ROC AUC |
|---|---|---|---|
| 1 | 0.8808 | 0.8831 | 0.9638 |
| 2 | 0.8807 | 0.8831 | 0.9637 |
| 3 | 0.8806 | 0.883 | 0.9637 |
| 4 | 0.8808 | 0.8831 | 0.9638 |
| 5 | 0.8806 | 0.883 | 0.9637 |
| Mean ± Std. dev | 0.8807 ± 0.0001 | 0.8830 ± 0.0001 | 0.9637 ± 0.0001 |

## IV. DISCUSSION

### A. Overall Traditional Model Performance

Linear and probabilistic models did not perform well. Gaussian Naïve Bayes, SGD SVM, and the Voting Classifier exceeded the baseline only marginally, with accuracy and F1 scores below 0.6.

Tree-based approaches demonstrated substantial improvements, with HistGradientBoosting reaching 68% accuracy. Random Forest results revealed relationship between tree depth and performance: while the constrained model (depth 8) had modest results similar to that of the linear models at 60% accuracy, allowing deeper trees (depth 30) produced a dramatic jump to 87% accuracy, suggesting that the linguistic feature space benefits from more complex decision boundaries.

### B. Ensemble Model Performance

The Bagging classifier achieved the best performance among traditional models, with 0.88 accuracy and 0.88 F1 score. Due to this strong performance on a traditional model, five-fold cross-validation was applied, showing stability with a standard deviation of ± 0.0001 across all metrics. The model consisted of 25 decision tree estimators, each trained on 60% of samples and 60% of features (69 features per tree).

*C. Neural Network/Deep Learning Performance*

Neural network architectures generally showed a capacity-performance relationship, with the smaller MLP achieving 74.3% accuracy while the expanded variant reached 80.3%, albeit with considerably extended training time. DistilBERT emerged as the strongest performer overall, achieving 90.3% accuracy, 90.3% F1 score, and 97.5% ROC AUC, though these gains came at the expense of substantially increased computational overhead during the fine-tuning process. This accuracy required over 14 hours of training time on a GeForce 4090, a significant slowdown compared to traditional methods.

*D. Overall Results*

These findings suggest that machine learning and deep learning approaches can reliably differentiate news quality using linguistic features extracted from link text. The effectiveness of ensemble methods proved particularly noteworthy with the Bagging classifiers having strong consistency over cross-validation folds (standard deviation ± 0.0001), suggesting generalization past our test conditions. Bagging methods excel in high-variance, high-dimensional settings by aggregating predictions from diverse estimators trained on random subsets of samples and features [9]. The 25 decision tree estimators were each trained on 60% of samples and 60% of features, enabling variance reduction while preserving underlying NLP patterns. This ensemble diversity suppressed noise within the large dataset, aligning with at least one previous work on ensemble methods for complex NLP features [3].

The neural models had a performance-efficiency trade-off. The small MLP accuracy was around 0.74, while the larger variant reached a 0.80 accuracy with substantially increased training time. Both MLPs outperformed linear models, suggesting non-linear decision boundaries.

SGD SVM and Gaussian Naïve Bayes appeared unable to capture the patterns in the 115-dimensional linguistic feature space. Random Forest performance was dependent on tree depth: shallow forests (depth 8) achieved near-baseline results, while deeper variants (depth 30) approached Bagging performance but required longer training times. This suggests that tree depth alone is not optimal.

The fine-tuned DistilBERT model, which was the only deep learning model tested here due to time/resource constraints, achieved the highest overall performance with 0.9027 accuracy, 0.9026 F1 score, and 0.9748 ROC AUC, but also the slowest train time.

These results show that ensemble methods like Bagging provide an effective balance for large-scale perceived news headline/link text quality classification, offering strong performance with low CPU-based training requirements. As expected, at least one deep learning model performed with superior accuracy, however, train time was greatly increased. Next steps could include hybrid approaches using BERT embeddings as features for ensemble models or to investigate dimensionality reduction techniques for improving efficiency without losing accuracy.

*E. Limitations*

There are limitations to this work. The PC1 scoring system, while having a high level of consensus from reputable sources [7], remains subject to human judgment, and some high PC1 scores can still contain bias [7]. The quality labels used here trade some accuracy for practical purposes as in the original work [7]. Some languages other than English were included in the dataset. Model optimization was constrained by computational resources and time limitations; we also were unable to train more modern variants of BERT due to time/resource constraints. Furthermore, while hyperparameter tuning was performed for key models, exhaustive grid search across all possible parameter combinations was not feasible. As a result the selected parameters may not represent best configurations for all models.

The linguistic feature set derived from the NLP detector included 115 features after sparsity filtering, but some features may not as directly relevant to link text quality assessment. Binary classification used a median threshold of 0.8163 and was selected to maximize data utilization while maintaining class balance. A multi-class or regression approach might provide more nuanced quality assessments but would require different evaluation and potentially larger datasets for each quality class. The domain-level labeling assumption treats all content from a given news organization as having uniform quality. While this approach enabled large-scale analysis, it does not account for potential quality variation within individual news sources or changes in editorial standards.

V. CONCLUSION

This analysis suggests that machine learning/deep learning approaches can effectively differentiate perceived lower-quality headline/link text from perceived higher-quality headline/link text on worldwide news pages. Using a dataset of 57 million headlines/links taken from news pages, both traditional ensemble methods with NLP features and deep learning approaches demonstrated good predictive capability. The CPU-based Bagging Classifier had the best performance (88.1% accuracy). This type of model also showed stability over cross-validation folds.

Deep learning, specifically fine-tuned DistilBERT was the most accurate with much longer train time. This result weighs the importance of modest accuracy improvements (~2%) against GPU usage. The poor performance of linear models and the success of tree-based and neural approaches suggest that news quality patterns based on these linguistic features are non-linear and complex. Traditional models like SGD SVM and Gaussian Naïve Bayes did not perform much better than the Dummy Classifier.

This analysis is one approach to news quality assessment using domain-level PC1 labels, linguistic features, and some deep learning. The stated limitations of quality labeling and the processing time for large-scale web data analysis, highlight the challenges inherent in this domain. Nevertheless, the effectiveness of ensemble methods and deep learning for news quality classification is promising.

Future opportunities exist to explore multi-class quality assessment beyond binary classification with additional exploration of the PC1 score. Additionally, the performance of additional deep learning models could be measured, including using generative AI to predict headline/link news quality or how generative AI output is classified by the models discussed in this work.

REFERENCES

[1] M. S. Jeong, L. Jacob A., and S. M. and Lavis, "The Viral Water Cooler: Talking About Political Satire

Promotes Further Political Discussion," *Mass Commun Soc*, vol. 26, no. 6, pp. 938–962, Nov. 2023, doi: 10.1080/15205436.2022.2138766.

[2] Fatemeh Torabi Asr and Maite Taboada, "Big Data and quality data for fake news and misinformation detection," *Big Data Soc*, vol. 6, no. 1, p. 2053951719843310, Jan. 2019, doi: 10.1177/2053951719843310.

[3] G. Gravanis, A. Vakali, K. Diamantaras, and P. Karadais, "Behind the cues: A benchmarking study for fake news detection," *Expert Syst Appl*, vol. 128, pp. 201–213, 2019, doi: https://doi.org/10.1016/j.eswa.2019.03.036.

[4] B. Probierz, P. Stefanski, and J. Kozak, "Rapid detection of fake news based on machine learning methods," in *Procedia Computer Science*, Elsevier B.V., 2021, pp. 2893–2902. doi: 10.1016/j.procs.2021.09.060.

[5] S. Rastogi and D. Bansal, "A review on fake news detection 3T's: typology, time of detection, taxonomies," *Int J Inf Secur*, vol. 22, no. 1, pp. 177–212, 2023, doi: 10.1007/s10207-022-00625-3.

[6] F.-J. Rodrigo-Ginés, J. Carrillo-de-Albornoz, and L. Plaza, "A systematic review on media bias detection: What is media bias, how it is expressed, and how to detect it," *Expert Syst Appl*, vol. 237, p. 121641, 2024, doi: https://doi.org/10.1016/j.eswa.2023.121641.

[7] H. Lin *et al.*, "High level of correspondence across different news domain quality rating sets," *PNAS Nexus*, vol. 2, no. 9, p. pgad286, Sep. 2023, doi: 10.1093/pnasnexus/pgad286.

[8] C. Brogly and C. McElroy, "Did ChatGPT or Copilot use alter the style of internet news headlines? A time series regression analysis," 2025. Accessed: Jun. 09, 2025. [Online]. Available: https://arxiv.org/pdf/2503.23811

[9] P. Bühlmann, "Bagging, Boosting and Ensemble Methods," *Handbook of Computational Statistics*, Jan. 2012, doi: 10.1007/978-3-642-21551-3_33.